\documentclass[letterpaper, 10 pt, conference]{IEEEtran}
\IEEEoverridecommandlockouts
\usepackage{cite}
\usepackage{amsmath,amssymb,amsfonts}
\usepackage{algorithmic}
\usepackage{graphicx}
\usepackage{textcomp}
\usepackage{xcolor}
\usepackage{array}
\usepackage{tikz}
\def\BibTeX{{\rm B\kern-.05em{\sc i\kern-.025em b}\kern-.08em
    T\kern-.1667em\lower.7ex\hbox{E}\kern-.125emX}}

\newcommand\copyrighttext{%
  \footnotesize \textcopyright 2023 IEEE. Personal use of this material is permitted.
  Permission from IEEE must be obtained for all other uses, in any current or future
  media, including reprinting/republishing this material for advertising or promotional
  purposes, creating new collective works, for resale or redistribution to servers or
  lists, or reuse of any copyrighted component of this work in other works.}
\newcommand\copyrightnotice{%
\begin{tikzpicture}[remember picture,overlay]
\node[anchor=south,yshift=10pt] at (current page.south) {\fbox{\parbox{\dimexpr\textwidth-\fboxsep-\fboxrule\relax}{\copyrighttext}}};
\end{tikzpicture}%
}

\title{Group Dynamics: Survey of Existing Multimodal Models and Considerations for Social Mediation\\
\thanks{$^{1}$Honda Research Institute USA, Inc. {\tt\smallskip \{hifza\_javed, njamali\}@honda-ri.com}}
}

\author{Hifza Javed$^{1}$, and Nawid Jamali$^{1}$}

\begin{document}
\maketitle
\copyrightnotice

\begin{abstract}
Social mediator robots facilitate human-human interactions by producing behavior strategies that positively influence how humans interact with each other in social settings. As robots for social mediation gain traction in the field of human-human-robot interaction, their ability to ``understand'' the humans in their environments becomes crucial. This objective requires models of human understanding that consider multiple humans in an interaction as a collective entity and represent the group dynamics that exist among its members. Group dynamics are defined as the influential actions, processes, and changes that occur within and between group interactants. Since an individual’s behavior may be deeply influenced by their interactions with other group members, the social dynamics existing within a group can influence the behaviors, attitudes, and opinions of each individual and the group as a whole. Therefore, models of group dynamics are critical for a social mediator robot to be effective in its role. In this paper, we survey existing models of group dynamics and categorize them into models of social dominance, affect, social cohesion, conflict resolution, and engagement. We highlight the multimodal features these models utilize, and emphasize the importance of capturing the interpersonal aspects of a social interaction. Finally, we make a case for models of relational affect as an approach that may be able to capture a representation of human-human interactions that can be useful for social mediation.
\end{abstract}


\section{Introduction}
\label{sec:intro}
Socially assistive robots have been leveraged in a number of application domains, including companion robots~\cite{odekerken2020mitigating}, robots for learning~\cite{belpaeme2018social}, and robots as assistants for older adults~\cite{niemela2019robots}. In all such applications, a key function of the robot is to act as a social mediator. In general, a social mediator facilitates human-human interactions with the goal of strengthening existing relationships between individuals and helping them form new connections. This requires that the robot produce behaviors that are able to impact how humans in its environment interact with one another. However, to produce behaviors that are relevant to the context of the social situation, the robot must first be able to ``understand'' the humans and the interaction dynamics among them in its environment.

Human-human interactions involve multiple individuals that interact with one another through complex verbal and nonverbal signals that are contextual and change over time. To successfully mediate such interactions, a robot must recognize these signals in real-time and understand what these mean in a given context. This understanding can then be leveraged to generate robot behaviors that can help build relationships, improve connectedness, and achieve group-specific goals.

Therefore, a prerequisite for social mediation is the understanding of group dynamics---the influential actions, processes, and changes that occur within and between group interactants~\cite{forsyth2018group}. Healthy group dynamics may manifest in various forms, including cooperation, creativity, cohesion, likeability, conflict resolution, open communication, and strong team performance. The interactions among group members have the potential to significantly impact an individual's behavior~\cite{lewin1951field}, and as such, the social dynamics present within a group can shape the attitudes, opinions, and behaviors of both individuals and the group as a whole~\cite{lee2014group}. Thus, the development of models of group dynamics is critical to ensure that social mediator robots can perform their roles effectively.

In this paper, we survey existing approaches to model group dynamics that are used in human-human-robot interaction (HHRI) studies and evaluate their ability to represent the interpersonal aspects of a social interaction. Specifically, we do this by collecting the features these models use to represent human-human interactions and analyzing their ability to capture the \textit{interplay} of behavioral cues like gaze, gestures, tone of voice, etc. We highlight some considerations for the next steps in this research field and make a case for models of relational affect that can be useful in building group-level understanding of human-human interactions.

\section{Existing modeling approaches}
\label{sec:existing-models}
Prior research has explored several different perspectives on modeling of group-level interaction phenomena. In this section, we describe the most frequently researched themes within group dynamics and their applications within the field of HHRI.
\subsubsection{\textbf{Social dominance}}  Social dominance is a relational, behavioral, and interacting state in which an individual achieves control or influence over others through communicative actions~\cite{strohkorb2015classification}. Studies have shown that highly dominant children receive more social attention than their peers, which can affect their performance in group learning environments~\cite{strohkorb2015classification}. It is also associated with harsh power tactics in workplace environments~\cite{aiello2018social}. Since social dominance is communicative in nature, it manifests in ways that make its detection, measurement, and classification easier compared to other phenomena. As a result, it has been investigated most frequently in existing research~\cite{strohkorb2015classification,jayagopi2009modeling,aran2010fusing}.

In HHRI literature, dominance is viewed from the perspective of imbalance in participation from the interactants. Therefore, dominance-related studies typically attempt to identify the least or most dominant individual in a group, in settings such as group meetings at the office~\cite{jayagopi2009modeling}, group project discussions among students~\cite{tennent2019micbot}, or interactive storytelling for child-robot interaction scenarios~\cite{strohkorb2015classification}.

\subsubsection{\textbf{Affect}} This category encompasses models based on a combination of affect, emotions, and mood. Affect is an umbrella term in psychology that refers to the experience of feelings, emotions, or moods~\cite{frijda1986emotions}. Emotions are short-term and intense whereas moods are long-term and diffuse affective states. Moods emphasize a stable affective context, while emotions emphasize affective responses to specific events \cite{xu2015mood}. Since affect is an internal feeling state, often without clear, homogeneous expressions, its evaluation can be a significant challenge.

Research related to affective models typically represent emotions on the valence-arousal circumplex~\cite{russell1980circumplex}, often utilizing learning methods to classify emotion states from facial expressions. Bottom-up approaches combine individual-level affective states to form a measure of group affect~\cite{vonikakis2016group}, whereas top-down approaches focus on using global, scene-level features~\cite{sharma2019automatic}. For a bottom-up approach, how the individual-level states may be combined to accurately capture group states is an important consideration.

\subsubsection{\textbf{Social cohesion}} Social cohesion is the bonding that affects the membership of an individual in a group and is the most important attribute of a successful group~\cite{carron1985development}. It may manifest in the form of interpersonal attraction between group members, commitment to the task of the group, susceptibilities to interpersonal influence, and loyalty to a group~\cite{friedkin2004social}. Groups with high levels of social cohesion among their members tend to be highly productive and successful, and the individual members experience high levels of personal satisfaction~\cite{loughead2016examination}. Manifestations of social cohesion also tend to be less observable than social dominance, which necessitates a reliance on post-interaction surveys to extract cohesion-related information~\cite{brawley1985development}.

Research related to computational modeling of social cohesion in groups may address certain social aspects of cohesion, such as interest level within a group~\cite{gatica2005detecting}, rapport~\cite{cassell2007coordination, gratch2006virtual}, attraction~\cite{madan2005voices}, or synchrony~\cite{campbell2008multimodal}, or may model cohesion as a whole, comprising of both task cohesion and social cohesion~\cite{hung2010estimating}.

\subsubsection{\textbf{Conflict resolution}} Conflict resolution is the process that two or more parties use to find a peaceful solution to their dispute. It can be important at the workplace where successful resolution of conflicts can lead to greater efficiency and goal achievement, and maintain a positive, comfortable environment for all employees. Since emotion regulation plays a key role in conflict resolution, some models of conflict resolution that target emotion regulation may be similar to models of affect~\cite{barsade2002ripple, jung2015using}. Emotion regulation is defined as a process by which individuals influence which emotions they have, when they have them, and how they experience and express them~\cite{costa2018regulating}.

In HHRI literature, conflict resolution is often studied within the context of child-play scenarios, where play behaviors are analyzed as constructive vs. non-constructive and social vs. non-social when conflicts related to object possession occur between the playmates~\cite{shen2018stop}. Prior studies have also established the effectiveness of a robot acting as an emotion regulator during group interactions, where mediative actions from the robot can influence conflict resolution skills~\cite{jung2015using}. These actions include, first, drawing attention to interpersonal conflicts arising from personal violations; second, discouraging interpersonal conflicts by identifying them as inappropriate; and finally, using humor to alleviate tension.

\subsubsection{\textbf{Engagement}} According to Sidner et al.~\cite{sidner2005explorations}, engagement is defined as “the process by which individuals involved in an interaction start, maintain and end their perceived connection to one another.” The definition of engagement and its specific observations may vary slightly with the interaction task and the experimental settings used in a study. For example, in a robot-mediated activity for older adults, engagement may be observed through an individual's attendance to the activity, the degree of attentiveness in the activity, the degree of active participation, the person’s attitude, and whether the person appears bored~\cite{fan2021sar}. On the other hand, for children with autism disorders, engagement may be determined as a combination of behaviors such as gaze focus, imitation, verbalizations, self-initiated interactions, triadic interactions, and smiling~\cite{javed2019robotic}. In general, however, engagement is understood to be related to interactivity, attention, and the general user experience~\cite{anzalone2015evaluating}.

\section{Multimodal features for modeling group dynamics}
\label{sec:multimodal-features}
Existing research has explored the use of multiple modalities of features to measure various aspects of group dynamics in HHRIs.  These modalities include audio, visual, and physiological measurements, alongside subjective measures derived from questionnaires and surveys. This section summarizes the commonly used features in models of group dynamics.

\subsubsection{\textbf{Audio features}} The use of audio cues to model group dynamics is an active area of research. Audio cues can provide valuable insights into how groups function and communicate, as well as to inform the design of more effective group communication strategies. Audio features consist of both verbal cues containing spoken content and vocal cues containing sounds, such as laughing or backchanneling~\cite{jung2013engaging,park2017telling}. Verbal and vocal cues are widely used in modeling group dynamics since verbal activity can be a reliable indicator of interactivity and communication between the interactants. While these cues may provide information about the states of the individuals, they can also capture the relational aspects of a human-human interaction with measures related to turn-taking behaviors and speaking times~\cite{smith2015real, strohkorb2015classification}. Table \ref{tab:audio-features} summarizes the key audio cues used to model group dynamics.

\begin{table}[htbp]
\caption{Key audio features for modeling group dynamics}
\label{tab:audio-features}
\begin{center}
\begin{tabular}{| m{0.1\textwidth} | m{0.2\textwidth}| m{0.1\textwidth} |}
\hline
\textbf{Feature} & \textbf{\textit{Description}}& \textbf{\textit{Model}}
\\
\hline
Total speaking energy & Speaker energy accumulated over the interaction & Dominance~\cite{jayagopi2009modeling}, Cohesion~\cite{hung2010estimating}\\
\hline
Total speaking length &  Total time that an individual speaks & Dominance~\cite{jayagopi2009modeling,skantze2017predicting,gillet2021robot,strohkorb2015classification}, Cohesion~\cite{hung2010estimating}, Engagement~\cite{fan2021sar}\\
\hline
Total speaking turns & Total number of  time intervals for which an individual's speaking status is active & Dominance~\cite{jayagopi2009modeling, tennent2019micbot,gillet2021robot}, Cohesion~\cite{hung2010estimating}\\
\hline
Backchannel length & Total number of short turns consisting of backchanneling behaviors & Cohesion~\cite{hung2010estimating}\\
\hline
Total number of interruptions &  Cumulative number of times that an individual starts talking while another speaker is active & Dominance~\cite{jayagopi2009modeling,skantze2017predicting,strohkorb2015classification}, Cohesion~\cite{hung2010estimating}\\
\hline
Total number of failed interruptions & Amount of time an individual talks but is unable to take over the floor from the speaker & Cohesion~\cite{hung2010estimating}\\
\hline
Total speaker floor grabs & Number of times an individual starts talking while there are other people speaking and all others stop talking before the individual does & Dominance~\cite{aran2010fusing}\\
\hline
Silence time & Amount of time spent in silence between exchanges in floor grabs & Cohesion~\cite{hung2010estimating}\\
\hline
Time between floor exchanges & Amount of time between all floor exchanges & Cohesion~\cite{hung2010estimating}\\
\hline
Speaking rate & Represents the pace of the conversation and is computed using the mrate estimator~\cite{morgan1997speech} & Cohesion~\cite{hung2010estimating}\\
\hline
Utterance addressee & Total time spent speaking to a certain addressee & Dominance~\cite{strohkorb2015classification}\\
\hline
Utterance type & Category or subject of each utterance from an individual & Dominance~\cite{strohkorb2015classification}\\
\hline
Participation unevenness & Difference between each individual's speech time and the average speech time for the group & Dominance~\cite{gillet2021robot}\\
\hline
Low-level descriptors of voice & Spectral and cepstral coefficients, voice quality, energy, logarithmic harmonic-to-noise ratio, spectral harmonicity, psychoacoustic spectral sharpness, etc. & Affect~\cite{sharma2019automatic}, Cohesion~\cite{sharma2019automatic}\\
\hline
Prosody & Prosodic features like pitch, loudness, and pace & Affect~\cite{sharma2019automatic}, Conflict (through emotion regulation)~\cite{costa2018regulating}\\
\hline
\end{tabular}
\end{center}
\end{table}

\subsubsection{\textbf{Visual features}} Visual cues provide valuable means of capturing interpersonal behaviors. Visual cues refer to any nonverbal signals that are produced by the members of the group, such as facial expressions, body language, and gestures. While it is generally easier to capture relational signals from audio cues, as they are often produced in the presence of other interactants, visual observations from a social interaction can also provide valuable information on the interplay of behaviors, such as gaze~\cite{jayagopi2009modeling}, changes in proximity between members~\cite{sharma2019automatic}, and the level of emotional expressions during an interaction~\cite{tarnowski2017emotion}. Table \ref{tab:visual-features} summarizes the key visual features used to model group dynamics.

\begin{table}[htbp]
\caption{Key visual features for modeling group dynamics}
\label{tab:visual-features}
\begin{center}
\begin{tabular}{| m{0.1\textwidth} | m{0.2\textwidth}| m{0.1\textwidth} |}
\hline
\textbf{Feature} & \textbf{\textit{Description}}& \textbf{\textit{Model}}
\\
\hline
Visual activity & A binary variable that indicates if a participant is visually active or inactive at each time step & Dominance~\cite{jayagopi2009modeling}\\
\hline
Level of subtle visual activity & Strength of visual activity extracted from close view cameras using residual coding bit rate & Cohesion~\cite{hung2010estimating}\\
\hline
Total visual activity length & The accumulated motion activity for an individual & Dominance~\cite{jayagopi2009modeling}\\
\hline
Total visual activity turns & Number of times an individual is continuously moving without breaks & Dominance~\cite{jayagopi2009modeling}\\
\hline
Turn duration & Total number of  time intervals for which an individual's visual activity status is active & Dominance~\cite{aran2010fusing}\\
\hline
Physical coercion & Total time spent engaging in physical coercion (shoving, grabbing, etc.) & Dominance~\cite{strohkorb2015classification}\\
\hline
Gaze focus & Total time spent looking at specified targets in the interaction such as the robot or other interactants & Dominance~\cite{strohkorb2015classification}, Cohesion~\cite{cassell2007coordination}, Engagement~\cite{sidner2005explorations, fan2021sar}\\
\hline
Head nods & Number of times an individual's head moves up and down in a single continuous movement on a vertical axis & Cohesion~\cite{cassell2007coordination}\\
\hline
Head pose & Head pose yaw angle & Engagement~\cite{fan2021sar,anzalone2015evaluating}\\
\hline
Sizes of face and body & Relative sizes of faces and bodies from image data & Affect~\cite{mou2015group,vonikakis2016group}\\
\hline
Relative location & Individual's distance from the group & Affect~\cite{mou2015group,vonikakis2016group}\\
\hline
Interaction time & Amount of time spent in the interaction & Engagement~\cite{sidner2005explorations}\\
\hline
Shared looking & Amount of time spent in coordinated gaze on targets of mutual interest & Engagement~\cite{sidner2005explorations}\\
\hline
\end{tabular}
\end{center}
\end{table}

\subsubsection{\textbf{Physiological features}} Physiological signals are generated by the body during the functioning of various physiological systems and refer to the bodily functions that can be measured and analyzed. These include signals that reflect changes to skeleton muscles (electromyogram or EMG), changes in heart beat or rhythm (electrocardiogram or ECG), changes in brain activity measured through the scalp (electroencephalogram or EEG), and changes in corneo-retinal potential measured from the human eye (electrooculogram or EOG)~\cite{rim2020deep}. Physiological signals can provide insights into the internal states of individuals, including their emotions~\cite{shu2018review}, stress levels~\cite{karthikeyan2011review}, and cognitive processes~\cite{haapalainen2010psycho}. By measuring physiological signals of individuals within a group, researchers can analyze how these signals correlate with the events of the interaction and influence the group dynamics. Table \ref{tab:psychological-features} summarizes the key physiological features used to model group dynamics.

\begin{table}[htbp]
\caption{Key physiological features for modeling group dynamics}
\label{tab:psychological-features}
\begin{center}
\begin{tabular}{| m{0.1\textwidth} | m{0.2\textwidth}| m{0.1\textwidth} |}
\hline
\textbf{Feature} & \textbf{\textit{Description}}& \textbf{\textit{Model}}
\\
\hline
Facial EMG & Facial electromyography (EMG) from the corrugator supercilii and zygomaticus major muscles & Affect~\cite{sawabe2022robot}\\
\hline
Electrodermal activity & Measurement of the electrical conductivity of the skin & Affect~\cite{sawabe2022robot,swangnetr2012emotional}\\
\hline
Heart rate & Mean and median heart rate as measures of centrality and standard deviation in heart rate & Affect~\cite{swangnetr2012emotional}, Conflict~\cite{costa2018regulating}\\
\hline
EEG & EEG signals collected from 14 different electrode channels  & Engagement~\cite{fan2021sar}\\
\hline
\end{tabular}
\end{center}
\end{table}

\subsubsection{\textbf{Subjective features}} In addition to the objective measures summarized above, some subjective measures have also been used for the modeling of group dynamics. These are typically obtained using pre- and post-experimental interviews and surveys to record the participants' self-ratings and feedback. 

Interview questions extracting cohesion-related information can be posed to the interactants  
and may extract information such as whether an interactant's partner listened to them, whether the partner annoyed them, and whether they would participate in the activity again with their partner~\cite{strohkorb2016improving}. Moreover, additional information such as the level of friendship and familiarity may also be obtained from interviews to contextualize the assessment of cohesion between interactants~\cite{strohkorb2016improving}. Measures of specific aspects of cohesion such as relationship satisfaction and trust can also be obtained by using a sub-scale of the Subjective Value Inventory~\cite{curhan2006people} to understand an interactant’s satisfaction of their relationship with their partner.

Subjective ratings of valence and arousal are also obtained from participants as ground truth labels of their affective states~\cite{sawabe2022robot,jung2016coupling}. Another commonly used method for extracting affective states is to elicit retrospective ratings of affect from the interactants themselves that are used to form a measure called group affective balance that can be used as a predictor of conflict resolution skills~\cite{jung2016coupling}.
    
\section{Discussion: considerations for social mediation}
While there is a wide body of qualitative evidence reporting the study of group dynamics, thorough quantitative analyses of these phenomena still remain a significant challenge. This may be attributed to difficulties with formal knowledge representation and the lack of data that can capture the nuances of human-human interactions in sufficient detail. Evaluating individual human states is a significant challenge and remains an active area of research. Therefore, it is inevitable that the study of multiple interactants and their dynamics presents additional complexities, which will have to be addressed by future research.

Various aspects of auditory signals have been investigated in the study of social dynamics within human-human interactions, as shown in Table~\ref{tab:audio-features}. The use of audio cues is not surprising, given that intentional vocal activity is typically expected to occur in the presence of other interactants. However, it must be noted that a deeper analysis of verbal behaviors, beyond the existence of vocal activity and turn-taking behavior, is not always available. Some studies perform an analysis to identify verbal behaviors based on its intent or meaning into categories such as acknowledgments, answers, influence, information request, rephrase, and signal of non-understanding~\cite{cassell2007coordination}. Such an approach provides valuable context about the nature of the interactions taking place and enable a thorough analysis of the group dynamics.

In comparison to audio features, it was found that the automated extraction of visual features was less common and instead, relied frequently on manual coding from human annotators~\cite{shen2018stop, strohkorb2016improving}. Moreover, deep learning-based approaches that work directly with image inputs are often used to process visual activity in group interactions~\cite{sharma2019automatic, tan2017group, guo2017group}, which also avoids the explicit extraction of visual features.

The use of physiological signals for internal state estimation in group interactions is limited in comparison to audio and visual modalities. Firstly, although many features have been tested, there is still no clear support for feature combinations that may be strongly correlated with affective state~\cite{shu2018review}. Additionally, individual differences in experiences of various affective states also complicates the use of a model utilizing physiological features~\cite{shu2018review}. However, unlike behavioral responses, some physiological variables are not under complete voluntary control~\cite{rasia2006there}, making the combination of physiological signals with other modalities valuable for internal human state estimation. 

Social dynamics prevailing in group interactions are represented not only by the emergence of certain behaviors in individuals but also by the interplay of behaviors such as gaze, gestures, tone and volume of voice, etc. Measures capturing this interplay of behaviors are able to represent the interpersonal aspects of the interaction. These can then be used to demonstrate not only the correlation between different modalities of behaviors but also the impact of this correlation on the evolving group dynamics. Some models surveyed in Section~\ref{sec:multimodal-features} use measures that represent such correlations between the audio and the visual behaviors. Strohkorb et al.~\cite{strohkorb2015classification} used combined behaviors like ``looking while listening'' and ``looking while talking'' to represent social dominance in groups of children as they interact with a robot, while Hung et al.~\cite{hung2010estimating} used features such as ``motion during overlapping speech'' and ``motion when not speaking'' to capture the correlations between body movements and speech.

This emphasis on capturing the interplay of behaviors that can represent interpersonal aspects of a social interaction gives rise to the concept of relational affect, which is a dyadic construct that represents affective states that an individual experiences from their interactions with others. According to Slaby~\cite{slaby2019relational}, relational affect ``does not refer to individual feeling states but to affective interactions in relational scenes, either between two or more interactants, or between an agent and aspects of [her or his] environment''. 
Relational affect is a consequence of the interaction itself in the form of an interplay of gaze, gesture, tone of voice etc.~\cite{slaby2019relational}, and it focuses on the observable expressions of affect between the interactants rather than the individual internal experiences of emotion.

Therefore, rather than viewing emotional expressions as indicators of internal state, these can instead be viewed as actions that can shape people’s relational orientation towards each other \cite{van2010interpersonal}. For example, expressions of anger or hostility can push people away, whereas expressions of joy or sadness can bring people together~\cite{jung2017affective}. These findings reiterate the need to add a third dimension of interpersonal orientation to the conventional valence-arousal model of affect, and support the proposal made by Jung \cite{jung2017affective} to represent interpersonal behaviors along an axis ranging from affiliative behaviors (behaviors that turn people towards each other) to distancing behaviors (behaviors that turn people away from each other).

Group-level models of human understanding that incorporate relational affect may be well-positioned for HHRI research, where they can be utilized by a social mediator robot to act in accordance with the social dynamics that exist within the group, thereby producing interaction strategies that are more effective not only in achieving the group's specific goals but also in building stronger relationships and improving overall connectedness between the group members.

\section{Conclusion}
A variety of multimodal features have been used in prior research to represent various aspects of social dynamics in human-human interactions. While most models of group dynamics capture the emergence of certain behaviors within individuals, some also capture the correlation between different modalities of behavioral indicators to represent certain nuances of interactions occurring between the individual interactants. We emphasize the need for capturing the interplay of behaviors between interactants that can represent the interpersonal aspects of a social interaction. We also make a case for models of relational affect, which can represent affective states that an individual experiences from their interactions with others. Such models can inform the interaction strategies produced by social mediator robot that aim to positively influence social dynamics prevailing within a group.
\newline

\bibliographystyle{IEEEtran}
\bibliography{main}

\end{document}